\documentclass[journal, twocolumn, 11 pt]{IEEEtran}
\newlength{\tocsep}
\usepackage[utf8]{inputenc}
\usepackage{verbatim}
\usepackage{url}

\usepackage{color}
\usepackage{multirow,array,graphicx}
\usepackage{amsmath}
\usepackage[linesnumbered,ruled,vlined,boxed]{algorithm2e}
\usepackage{multirow}
\usepackage{placeins}

\usepackage{booktabs} 
\usepackage{amsmath}

\usepackage{caption,subcaption}
\usepackage{tikz}
\usetikzlibrary{positioning,fit}
\usetikzlibrary{matrix}

\definecolor{darkgreen}{rgb}{0.0, 0.2, 0.13}
\definecolor{darkolivegreen}{rgb}{0.33, 0.42, 0.18}

\newcommand{\ie}{\textit{i.e.,}\ }

\author{
Leo Cazenille$^{1}$

{1} Department of Information Sciences, Ochanomizu University, Tokyo, Japan
}

\title{Comparing reliability of grid-based Quality-Diversity algorithms using artificial landscapes}

\begin{document}

\flushbottom
\maketitle
\thispagestyle{empty}

\begin{abstract}
Quality-Diversity (QD) algorithms are a recent type of optimisation methods that search for a collection of both diverse and high performing solutions. They can be used to effectively explore a target problem according to features defined by the user.
However, the field of QD still does not possess extensive methodologies and reference benchmarks to compare these algorithms.
We propose a simple benchmark to compare the reliability of QD algorithms by optimising the Rastrigin function, an artificial landscape function often used to test global optimisation methods.
\end{abstract}

\begin{IEEEkeywords}
Quality-Diversity, Unconstrained Optimisation, Exploration, MAP-Elites
\end{IEEEkeywords}

\section{Introduction}
Search optimisation algorithms are popular methods to automatically explore a search space to find high-performing solutions. In such cases, the goal is traditionally to find the single best solution. However, in some problems it is useful for the search process to explore a range of both diverse and high-performing solutions.
This approach is realised by a recent family of optimisation algorithms named Quality-Diversity (QD) algorithms or Illumination algorithms~\cite{pugh2016quality,cully2018quality}, such as Map-Elites~\cite{mouret2015illuminating} and Novelty Search with Local Competition~\cite{lehman2011evolving}.
Algorithms like Map-Elites~\cite{mouret2015illuminating} are grid-based, and regroup the explored solutions in a grid of elites. This produces sets of high-performing solutions that vary according to features defined by the user, represented as axes of the grid.

These algorithms are particularly successful in evolutionary robotics problems~\cite{mouret2015illuminating,cully2015robots,duarte2018evolution}, either by improving diversity to overcome deceptive search spaces~\cite{lehman2013effective}, or by identifying and exploiting the generated repertoire of solutions~\cite{mouret2015illuminating}.

QD algorithms are attractive methods to explore features landscapes of difficult and ill-defined problems, as the diversification of the explored solutions could help a traditional optimisation process to cope with highly multi-modal and deceptive target functions and prevent it from getting stuck in local-optima.

Here, we propose a method to compare grid-based QD algorithms by exploring artificial landscapes functions often used to compare global optimisation algorithms~\cite{dieterich2012empirical}. The goal is not just to find the global minimum as it is traditionally the case, but to exhaustively explore the landscape of this function.
In this context, we compare the reliability~\cite{mouret2015illuminating} (\ie how close the solutions found for each bin of the grid are to the oracle values) of two versions of the MAP-Elites algorithms on the Rastrigin function~\cite{rastrigin1974systems}.

\section{Methods}

\subsection{Benchmark}
\begin{figure}[h]
\begin{center}
\includegraphics[width=0.38\textwidth]{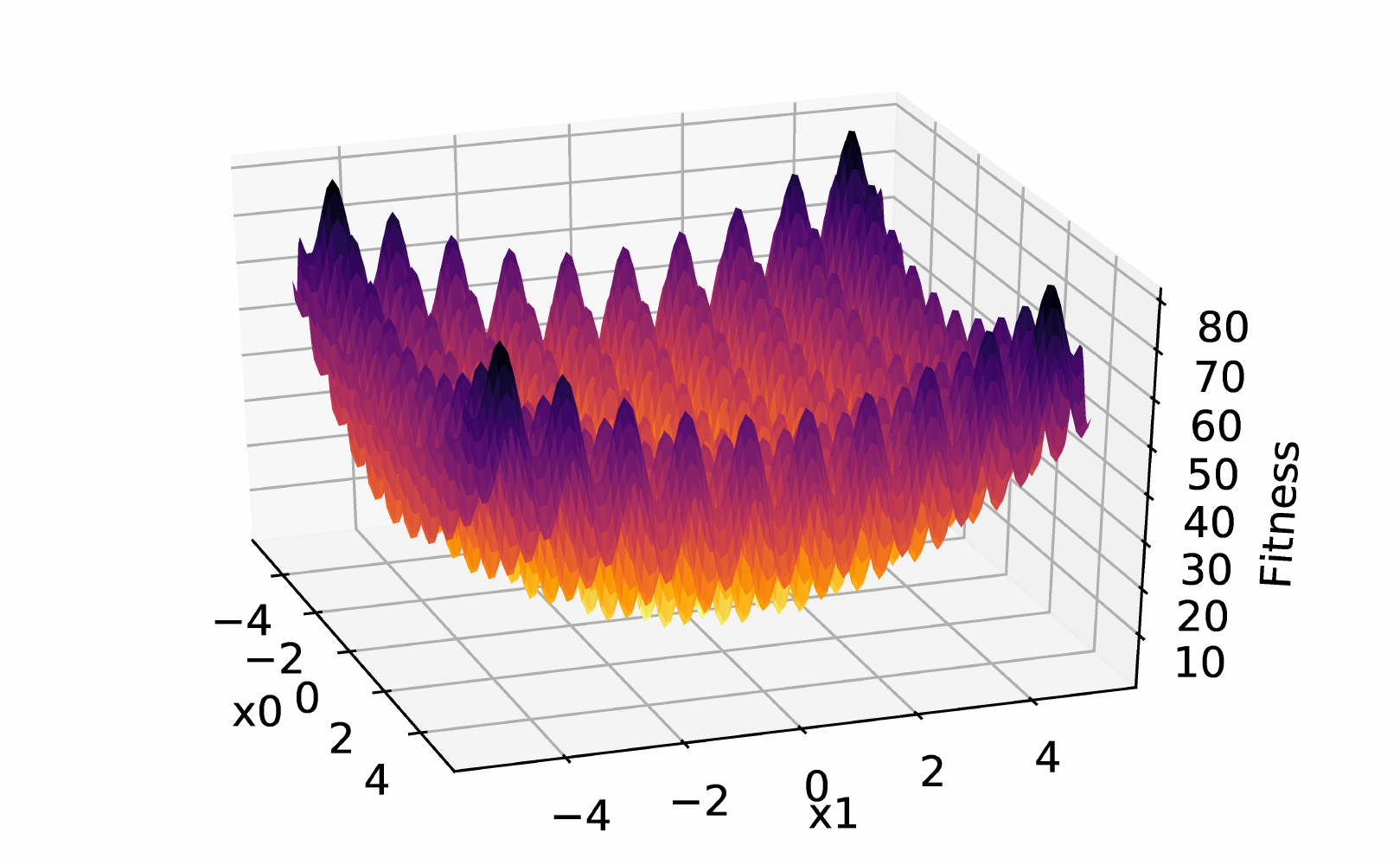}
\caption{\label{fig:refRastrigin} Surface plot of the Rastrigin function~\cite{rastrigin1974systems} with 2 dimensions. Global maxima are close to the corners and the single global minimum is at $x_i = 0$ where $f(x)=0$.}
\end{center}
\end{figure}

We minimise the Rastrigin~\cite{rastrigin1974systems} function (Fig.~\ref{fig:refRastrigin}). It is often used as a performance test problem for single-objective optimisation algorithms. On an $n$-dimensional domain~\cite{muhlenbein1991parallel}, it is defined by:
\begin{equation}
    f_n(\textbf{x}) = A n + \sum_{i=1}^n (x_i^2 - A cos(2\pi x_i))
\end{equation}
where $A = 10$ and $x_i \in [-5.12, 5.12]$. Finding the minimum of this function is difficult due to its large search space and high multimodality.
It is easy to increase the difficulty of this benchmark by increasing the dimensionality $n$ of the target function.

We consider two features to illuminate, corresponding to the first two components of $\textbf{x}$: $x_0$ and $x_1$.

\subsection{Reliability Metric}
As defined in~\cite{mouret2015illuminating}, the reliability of a QD algorithm corresponds to how close the solutions found for each bin of the grid are to the oracle values. Reliability can be measured locally (at the level of a single bin) or globally (covering the entire grid).
We create a reference grid $M$ of the oracle values by running the MAP-Elites~\cite{mouret2015illuminating} algorithm to illuminate the Rastrigin~\cite{rastrigin1974systems} function on a 2-dimensional domain.

The bin of $M$ at position $x,y$ is termed $M_{x,y}$. We set $M_{\text{max}}$ as the maximal quality values in $M$, and $n(M)$ as the number of filled bins in $M$.
The local reliability $L(m_{x,y})$ of bin $m_{x,y}$ for grid $m$ is:
\begin{equation}
    L(m_{x,y}) = \begin{cases}
        0, & \text{if either $m_{x,y}$ or $M_{x,y}$ is not filled}.\\
        E(m_{x,y}), & \text{otherwise}.\\
    \end{cases}
\end{equation}
\begin{equation}
E(m_{x,y}) = 
        \text{max}\left(
        \frac{M_{\text{max}} - m_{x,y}}
            {M_{\text{max}} - M_{x,y}}
        , 0\right)
\end{equation}

The global reliability $G(m)$ for grid $m$ is defined as:
\begin{equation}
    G(m) = \frac{1}{n(M)} \sum_{x,y} L(m_{x,y})
\end{equation}

All benchmarks were conduced using the QDpy~\cite{qdpy} framework.

\section{Results}
We use the proposed methodology to compare two versions of the Map-Elites~\cite{mouret2015illuminating} algorithm according to their global reliability.
\textbf{ME1} uses polynomial bounded mutations, like the original Map-Elites algorithm and as described in~\cite{deb2002fast} (mutation prob.: $0.5$, eta: $10$).
\textbf{ME2} uses Gaussian mutations (mutation prob.: $0.5$, mean: $0$, stddev: $1.0$).
Both versions use $64$ bins per feature (\ie $4096$ bins per grid). This number of bins is chosen arbitrarily to illustrate our methodology.

We define the \textbf{Reference} case as an oracle that describes the illuminated grid of solutions for the 2-dimension Rastrigin function. Other methods are then compared to the \textbf{Reference} case. It is possible to compare illuminations of $f_k$ to illuminations of $f_l$ with $k\ge l$ because the Rastrigin function~\cite{rastrigin1974systems} is separable.

\begin{figure}[h]
\centering

\includegraphics[width=0.45\textwidth]{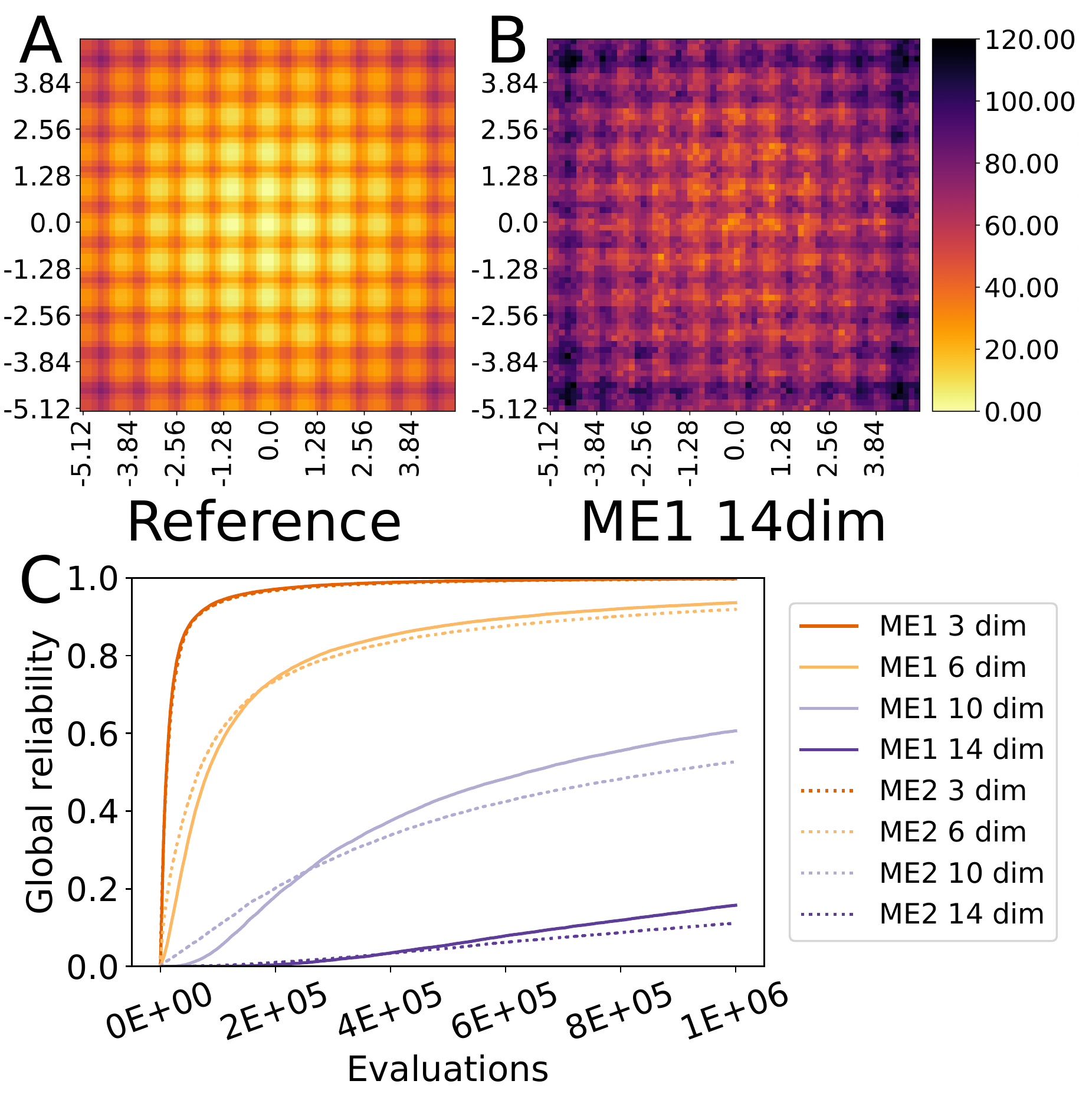}

\caption{A, B: Quality of each bin of the final grids. C: Global reliability of the ME1 and ME2 algorithms on the Rastrigin function with $N=3,6,10,14$ dimensions compared to the reference case of Rastrigin with $2$ dimensions.}
\label{fig:results}
\end{figure}

Figure~\ref{fig:results} compares the results obtained after $1000000$ evaluations, for both the \textbf{ME1} and \textbf{ME2} methods used to illuminate the Rastrigin function with varying number of dimensions. Figures~\ref{fig:results}A and B show examples of the final grids for the reference case and for \textbf{ME1} applied on the 14-dimension Rastrigin function.
Figure~\ref{fig:results}C presents the evolution of global reliability by evaluations, for all tested cases.

\section{Conclusion}

We presented a simple method to compare the reliability of grid-based QD algorithms with artificial landscapes functions. Namely, we compared two version of the Map-Elites~\cite{mouret2015illuminating} algorithm on the Rastrigin function~\cite{rastrigin1974systems} often used to compare global optimisers~\cite{dieterich2012empirical}.

We plan to extend this study to other artificial landscapes functions for unconstrained or constrained optimisation.
Our methodology could be extended to test algorithms with grids composed of several different numbers of bins.

\FloatBarrier

\begin{thebibliography}{10}

\bibitem{pugh2016quality}
Justin~K Pugh, Lisa~B Soros, and Kenneth~O Stanley.
\newblock Quality diversity: A new frontier for evolutionary computation.
\newblock {\em Frontiers in Robotics and AI}, 3:40, 2016.

\bibitem{cully2018quality}
Antoine Cully and Yiannis Demiris.
\newblock Quality and diversity optimization: A unifying modular framework.
\newblock {\em IEEE Transactions on Evolutionary Computation}, 22(2):245--259,
  2018.

\bibitem{mouret2015illuminating}
Jean-Baptiste Mouret and Jeff Clune.
\newblock Illuminating search spaces by mapping elites.
\newblock {\em arXiv preprint arXiv:1504.04909}, 2015.

\bibitem{lehman2011evolving}
Joel Lehman and Kenneth~O Stanley.
\newblock Evolving a diversity of virtual creatures through novelty search and
  local competition.
\newblock In {\em Proceedings of the 13th annual conference on Genetic and
  evolutionary computation}, pages 211--218. ACM, 2011.

\bibitem{cully2015robots}
Antoine Cully, Jeff Clune, Danesh Tarapore, and Jean-Baptiste Mouret.
\newblock Robots that can adapt like animals.
\newblock {\em Nature}, 521(7553):503, 2015.

\bibitem{duarte2018evolution}
Miguel Duarte, Jorge Gomes, Sancho~Moura Oliveira, and Anders~Lyhne
  Christensen.
\newblock Evolution of repertoire-based control for robots with complex
  locomotor systems.
\newblock {\em IEEE Transactions on Evolutionary Computation}, 22(2):314--328,
  2018.

\bibitem{lehman2013effective}
Joel Lehman, Kenneth~O Stanley, and Risto Miikkulainen.
\newblock Effective diversity maintenance in deceptive domains.
\newblock In {\em Proceedings of the 15th annual conference on Genetic and
  evolutionary computation}, pages 215--222. ACM, 2013.

\bibitem{dieterich2012empirical}
Johannes~M Dieterich and Bernd Hartke.
\newblock Empirical review of standard benchmark functions using evolutionary
  global optimization.
\newblock {\em arXiv preprint arXiv:1207.4318}, 2012.

\bibitem{rastrigin1974systems}
Leonard~Andreevi{\v{c}} Rastrigin.
\newblock Systems of extremal control.
\newblock {\em Nauka}, 1974.

\bibitem{muhlenbein1991parallel}
Heinz M{\"u}hlenbein, M~Schomisch, and Joachim Born.
\newblock The parallel genetic algorithm as function optimizer.
\newblock {\em Parallel computing}, 17(6-7):619--632, 1991.

\bibitem{qdpy}
L.~Cazenille.
\newblock Qdpy: A python framework for quality-diversity.
\newblock \url{https://gitlab.com/leo.cazenille/qdpy}, 2018.

\bibitem{deb2002fast}
Kalyanmoy Deb, Amrit Pratap, Sameer Agarwal, and TAMT Meyarivan.
\newblock A fast and elitist multiobjective genetic algorithm: Nsga-ii.
\newblock {\em IEEE transactions on evolutionary computation}, 6(2):182--197,
  2002.

\end{thebibliography}

\end{document}